\documentclass[a4paper,twoside]{article}

\usepackage{epsfig}
\usepackage{subfigure}
\usepackage{calc}
\usepackage{amssymb}
\usepackage{amstext}
\usepackage{amsmath}
\usepackage{amsthm}
\usepackage{multicol}
\usepackage{pslatex}
\usepackage{apalike}
\usepackage{SCITEPRESS}     

\usepackage{times}
\usepackage{epsfig}
\usepackage{graphicx}
\usepackage{amsmath}
\usepackage{amssymb}

\usepackage{algorithmicx}
\usepackage{algpseudocode}
\usepackage{algorithm}

\def\rot{\rotatebox}
\usepackage{multirow}

\begin{document}

\title{Depth from Small Motion using Rank-1 Initialization}

\author{\authorname{Peter O. Fasogbon}
\affiliation{Nokia Technologies, 33100 Tampere, Finland}
\email{ \{peter.fasogbon\}@nokia.com }
}


\abstract{Depth from Small Motion (DfSM) \cite{Hyowon:2016} is particularly interesting for commercial handheld devices because it allows the possibility to get depth information with minimal user effort and cooperation. Due to speed and memory issue on these devices, the self calibration optimization of the method using Bundle Adjustment (BA) need as little as 10-15 images. Therefore, the optimization tends to take many iterations to converge or may not converge at all in some cases. This work propose a robust initialization for the bundle adjustment using the rank-1 factorization method \cite{Tomasi1992}, \cite{Aguiar:1999}.  We create a constraint matrix that is rank-1 in a noiseless situation, then use SVD to compute the inverse depth values and the camera motion. We only need about quarter fraction of the bundle adjustment iteration to converge. We also propose grided feature extraction technique so that only important and small features are tracked all over the image frames. This also ensure speedup in the full execution time on the mobile device. For the experiments, we have documented the execution time with the proposed Rank-1 initialization on two mobile device platforms using optimized accelerations with CPU-GPU co-processing. The combination of Rank 1-BA generates more robust depth-map and is significantly faster than using BA alone. 		
}

\onecolumn \maketitle \normalsize \vfill

\section{\uppercase{Introduction}}
\label{sec:introduction}

	The use of smartphones is growing continuously nowadays and the level of expectation around what these cameras can do is increasing year by year. Mobile consumers are starting to expect more technological capabilities from visual applications on their mobile devices. These applications include but not limited to camera refocusing, 3D parallax, augmented reality and extended depth of field \cite{Barron:2015}. To meet these needs, estimating three dimensional information is becoming an increasingly important technique, and numerous research efforts have focused on computing accurate three dimensional information at a low cost, without the need for additional devices or camera modifications. One research direction that has recently led to renewed interest is the depth estimation from image sequences acquired from narrow/small baseline in the range of about 8$mm$. 	This is popularly known as Depth from Small Motion (DfSM), and many research contributions have been made over the years \cite{Gallup:2014}, \cite{micro-baseline-stereo}, \cite{Hyowon:2016}, \cite{Corcoran:2017}, \cite{Ham:2017}. For hand-held cameras, small amount of motion is always present, which can occur accidentally as a result of hand-shaking motion, tremble, source vibration etc. Depth-map generation using these small motions can be offered to consumers to accompany their selfies, bothie and portraits camera shots. 

\subsection{Background}
These DfSM methods all have their base on the popular Structure from Motion (SfM) \cite{Schonberger:2016} and Multi-View Stereo (MVS) \cite{Seitz:2006} techniques. The SfM techniques assume that a good two-view reconstruction can be obtained with algebraic methods, which in turn depend on adequate baseline between overlapping views. The baseline between sequences of frames captured as a sudden motion in DfSM is considered so small which restricts the viewing angle of a three-dimensional point to less than 0.2$^\circ$ \cite{Gallup:2014}. Due to this limitation, the popular SfM method fails \cite{Koenderink:91}, \cite{Schonberger:2016} and the computed depth-map will be highly penalized. Bundle adjustment (BA) \cite{Hartley:2003:MVG:861369}, \cite{Triggs:1999:BAM:646271.685629} is an indispensable procedure in the SFM, and use a basic cost function to evaluate the reprojection error from {\bf U}ndistorted to {\bf D}istorted image domain with non-linear least square. This is used to iteratively refine the camera parameters and three-dimensional points required to generate the depth-map. The bundle adjustment used for SfM methods are not suitable for small motions, therefore a modified bundle adjustment is proposed in \cite{Hyowon:2016} under inverse depth representation. In this case, the reprojection error is estimated from mapping the points in the {\bf D}istorted to {\bf U}ndistorted domain.  The sparse three-dimensional points are created by random depth initialization \cite{Gallup:2014}, then plane sweeping based image matching \cite{Collins:1996} is employed to create the depth-map. Finally, Markov Random Field \cite{Komodakis:2009} approach is employed to regularize the estimated depth-map effectively.




\subsection{Problem Statement}		
Although DfSM algorithm is specially designed for small baselines, the estimated camera poses become unreliable if the motion is unreasonably small. It is assumed that the required minimum baseline to apply this approach is reasonable when large number of frames are acquired, approximately 30 frames  \cite{Hyowon:2016}. As a result of limited memory space on mobile devices and the execution time issue, we are restricted to only use 10-15 frames in the depth-map generation on the mobile devices. The consequence of using this small number of frame means that the self calibrating bundle adjustment may not converge fast enough or not converge at all. In addition, as a result of lack of features near the image border, the estimated radial distortion parameters diverge beyond their bound and may not give meaningful estimation.

\vspace{0.5cm}

Due to these problem, the BA do not always give correct estimates of the camera parameters and inverse depth values. One solution to tackle this issue might be to use very high number of feature points in the order of 10,000 and above, or include an additional photometric bundle adjustment \cite{AlismailBL16b} step if one is restricted to small number of feature points. In addition, one can bound the camera parameters during the optimization. However, these solutions only introduce additional complexity to the system optimization. Therefore, a good initialization for the bundle adjustment is vital for the depth-map accuracy, so we proposed to use factorization technique based on Rank-1 suitable for inverse depth representation.

\subsection{Summary}
In this paper, we describe an uncalibrated Depth from Small Motion technique using rank-1 initialization. This approach provides a better initialization for the bundle adjustment procedure that takes too much or doesn't converge under DfSM. This is particularly suitable and targeted to speedup processes for the deployment of the DfSM algorithm on consumer smartphone devices. The Rank-1 factorization does not only speed up the convergence process but also allow good initialization for accurate depth-map generation. Thanks to rank-1 initialization, self calibrating Bundle Adjustment (BA)  is able to converge in as little as 10-20 iterations with 10 images. We also proposed a grided feature extraction to speedup feature tracking process of the algorithm. Finally, we optimized various parts of the original algorithm \cite{Hyowon:2016} using GPU OpenCL and other CPU multi-threading techniques. This makes it possible to produce a detail experiment on a mobile device under ANDROID platform. 

\vspace{0.5cm}
In the next section, we present the uncalibrated rank-1 factorization for the DfSM problem. Experiments and performance evaluation with the proposed method as compared to optimized CPU only implementations are provided in section 3.  Finally, we made conclusion and future direction in this work.


\begin{figure}
	\begin{center}
		\includegraphics[height=2.9cm, width=7.5cm]{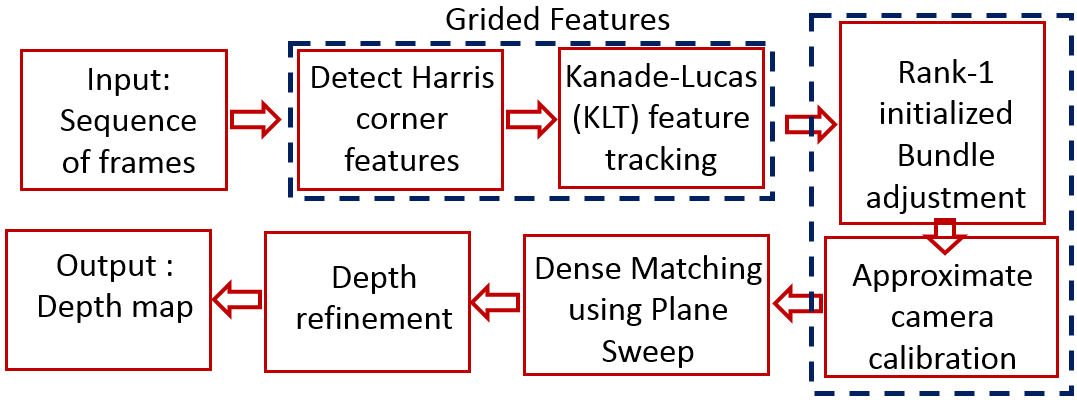}
	\end{center}
	\caption{DfSM framework. Our main technical contributions are in the dashed enclosed boxes.
	}
	\label{fig:onecol1r}
\end{figure}

\section{\uppercase{DfsM with Rank-1 Initialization}}
Fig. \ref{fig:onecol1r} illustrates the general overview of the DfSM algorithm for the depth-map generation in this work. Some consistent good features over all the video frames were extracted using grided feature tracking approach proposed in this work. Then, we initialize the bundle adjustment procedure using the rank-1 factorization technique, the outcome are the optimized camera parameters and the inverse depth point values. Finally, the estimated inverse depth point values and the camera parameters are used under a dense matching method to create the depth-map. In the following part of the section, we start first with coordinate representation used in this paper, then we explain the proposed grided-feature extraction, and Rank-1 initialization methods. In the final part of the section, we summarize the  DfSM algorithm.


\vspace{1cm}

\begin{figure}
	\begin{center}
		\includegraphics[height=3.9cm, width=7.5cm]{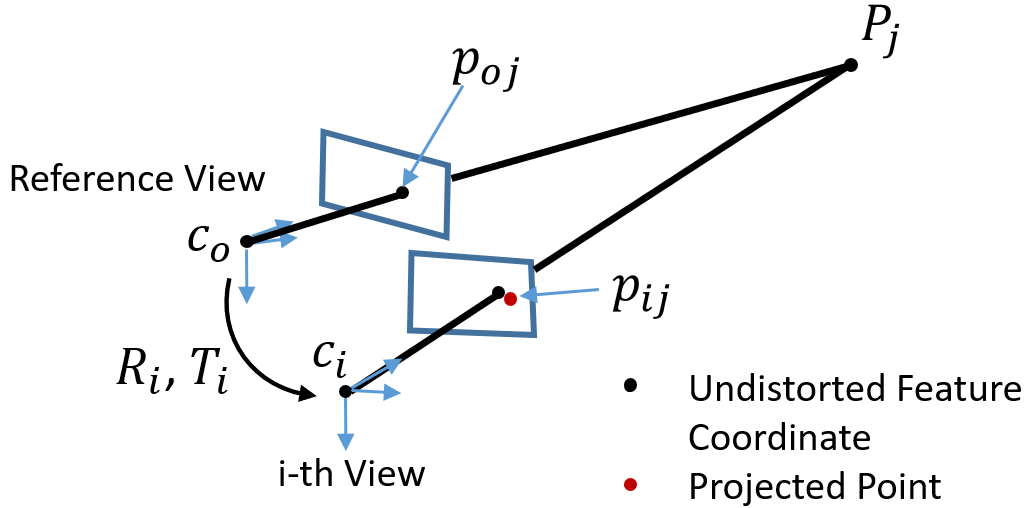}
	\end{center}
	\caption{Small motion geometry used in DfSM for uncalibrated camera using undistorted coordinates (i.e assume no lens distortion). 
	}
	\label{fig:onecol1}
\end{figure}

\noindent
{\bf Coordinate Representation}\\
Fig. \ref{fig:onecol1} illustrates the reference view coordinate origin $C_0$ with undistorted pixel $p_{0j}(u,v)$. The back-projection of $p_{0j}(u,v)$  onto both the 3D coordinate and $i$-th view with coordinate origin $C_i$ is denoted as $P_{j}$ and $p_{ij}$. Both $i$ and $j$ signifies ${j=1\ldots m}$ points and ${i=1\ldots n}$ views.  The $i$-th camera is related to the reference plane by rotation matrix $R_i$ followed by translation  $T_i$.  The backprojected 3D point can be parametrized using the inverse depth $\omega_j$ as shown in equation (\ref{fig:s}), where $(x_j , y_j)$ is the normalized coordinate of $p_{0j}(u,v)$ derived from using the inverse of the intrinsic camera matrix $K$\cite{Hyowon:2016} \cite{Hartley:2003:MVG:861369} that embeds both the focal length and principal point. 

\begin{equation}  
\begin{split}
P_j = \frac{1}{\omega_j}[x_j, y_j, 1]^T
\end{split}
\label{fig:s}
\end{equation}  

Note that the earlier expressions and explanations assume no lens distortion whatsoever. Indeed, lenses are affected by distortion and the most common one is the radial distortion  \cite{Hartley:2003:MVG:861369}. To remove these radial distortion, we need to deduce a mapping functions $\mathcal{F}$ proposed in \cite{Hyowon:2016} with radial coefficients $k_1, k_2$ . This function $\mathcal{F}$ helps to map the distorted points $\tilde{p}_{0j}, \tilde{p}_{ij}$ to undistorted ones $p_{0j}, p_{ij}$ respectively using iterative inverse mapping. For the simplicity of the rest of this section, we assume the radial lens distortions have been removed.




\subsection{Grided Feature Tracking}	
The feature extraction step is an interest for us here as a means to speed up the execution time of the whole algorithm. Our main goal is to reduce as much as possible the total number of feature points that is tracked all along the frame sequences. We proposed what is called a grided feature extraction approach.  The full resolution image is first divided into grids of fixed sizes. Then, we proceed by extracting only strongest harris corners \cite{Harris88alvey} in the enclosed grids. We use Shi-Tomasi score as the measure of best feature in an enclosed grid \cite{ShiTomasi:1994}. The correspondence feature locations to the other frames are found by Kanade-Lukas-Tomashi (KLT) method \cite{Lucas:1981}. 



\subsection{Initialization using Rank-1 }
\begin{figure}
	\begin{center}
		\includegraphics[width=0.5\linewidth]{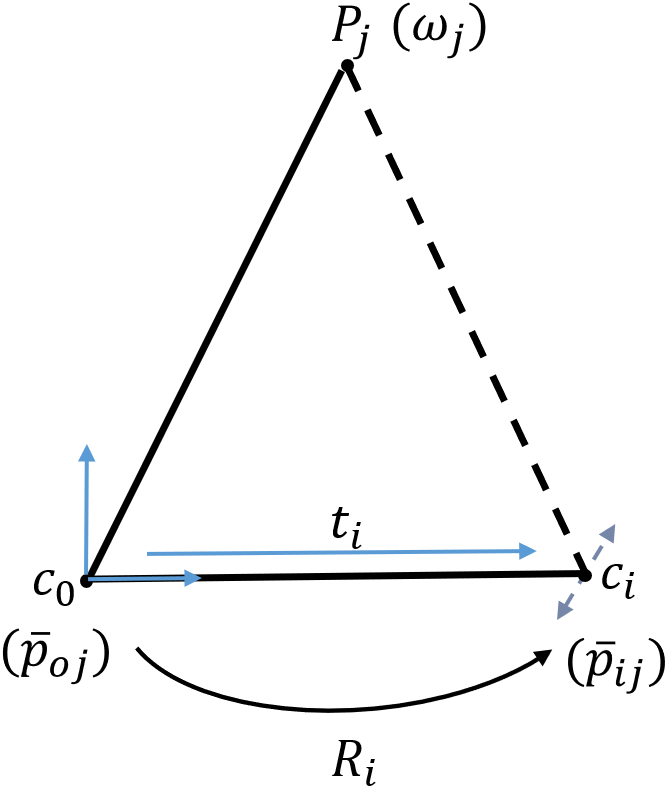}
	\end{center}
	\caption{The setup constraint for the rank-1 problem represented in normalized image coordinate $\bar{p}$ from the reference plane.
	}
	\label{fig:onecol2}
\end{figure}

Without abuse of notation, we denote the normalized coordinate of both $p_{0j}$ and $p_{ij}$ in the reference coordinate as $\bar{p}_{0j}(x,y)$ and $\bar{p}_{ij}(x,y)$ respectively. Fig. \ref{fig:onecol2} illustrates the coordinate origin that have been centered on the reference plane. In a perfect case depicted by the figure, we can see that $c_0$ belonging to the reference plane is fixed at origin, and the optical axis formed by $c_i$ is parallel to that of $c_0$.  

\vspace{0.3cm}

We can determine the relative camera rotation $R_i$ between keyframe $c_0$ and frame $c_i$ by \cite{Kneip:2012}. Given these rotations and set of corresponding features $\bar{p}_{0j} \leftrightarrow \bar{p}_{ij}$, we can estimate an optimal translation $t_i$ between $c_0$ and $c_i$ using factorization method.  

We create a form of flow representation between the origin $c_0$ and $c_i$ using the inverse depth point representation $P_j$ as a constraint for the factorization problem. We formulate the transformation of point $\bar{p}_{ij}$ located on i-th plane onto the reference coordinate as $\mathbf{p}_{ij}$ which is computed as follows:

\begin{equation}
\mathbf{p}_{ij} = - R_i~ \bar{p}_{ij}
\label{eqn:transh}
\end{equation}

\vspace{0.5cm}

\noindent
In fig. \ref{fig:onecol2}, we have only shown that the position of $c_i$ is only approximated ($c_i \approx - R_i~t_i$). This means that $c_i$ and $\omega_j$ are the most penalized. Therefore, with factorization method one should be able to determine the optimal value for these parameters even under noise perturbations.

\vspace{0.5cm}
By analysing the inverse depth representation in equation (\ref{fig:s}), one can see that $(x_j, y_j)$ are known from the position of the features in the reference frame, so we only need to solve for the inverse depth value $\omega_j$. The problem to solve is represented in equation (\ref{err:rep4}), where the Left-Hand-Side matrix $\mathbb{M}(3n \times m)$ is made up of flow representation $\mathbf{p}_{ij}$ to create a significantly over-constrained system of equations. The factorization  of $\mathbb{M}$ should give the Right-Hand-Side which consist of the translation matrix $\mathbb{C}(3n \times 1)$ and inverse depth matrix $\mathbb{D}(1 \times m)$.

\begin{equation}  
\begin{split}
\mathbb{M} = \mathbb{C} ~\mathbb{D}^T\\
\begin{bmatrix}
\mathbf{p}_{11} & \ldots & \mathbf{p}_{1m}\\
& \vdots & \\
\mathbf{p}_{n1} & \ldots & \mathbf{p}_{nm}\\
\end{bmatrix} =  \begin{bmatrix}
c_{1} \\
\vdots \\
c_{n}\\
\end{bmatrix}  \begin{bmatrix}
d_{1}  & \ldots & d_m \\
\end{bmatrix}
\end{split}
\label{err:rep4}
\end{equation}
\vspace{1cm}

The factorization problem in equation (\ref{err:rep4}) has been reduced to rank-1 problem, thanks to the inverse depth representation which means only the inverse depth is determined. This rank-1 factorization is extensively studied in computer vision community over the years \cite{Tomasi1992}, \cite{micro-baseline-stereo}, \cite{Aguiar:1999}, \cite{Aguiar:1999:2},\cite{ChengzhouTang:2017}.

The solution is formulated as a form of non-linear optimization in equation (\ref{err:rep}) and solved using SVD \cite{Golub:1996:MC:248979}. From the equation, $\|\mathbf{p}_{i,j} - c_{i}\omega_{j}\|^2$ is a geometric error to be minimized between the flow representation $\mathbf{p}_{i,j}$  and the estimated parameters ($c_{i},\omega_{j}$).

\begin{equation}  
\begin{split}
\mathbb{M} - \mathbb{C} ~\mathbb{D}^T\\
{\mathrm{minimize}}~~\sum\limits_{i=1}^{n}\sum\limits_{j=1}^{m}\|\mathbf{p}_{ij} - c_{i}\omega_{j} \|^2\\
\end{split}
\label{err:rep}
\end{equation}

\begin{figure}
	\begin{center}
		\includegraphics[height=4.1cm, width=7.5cm]{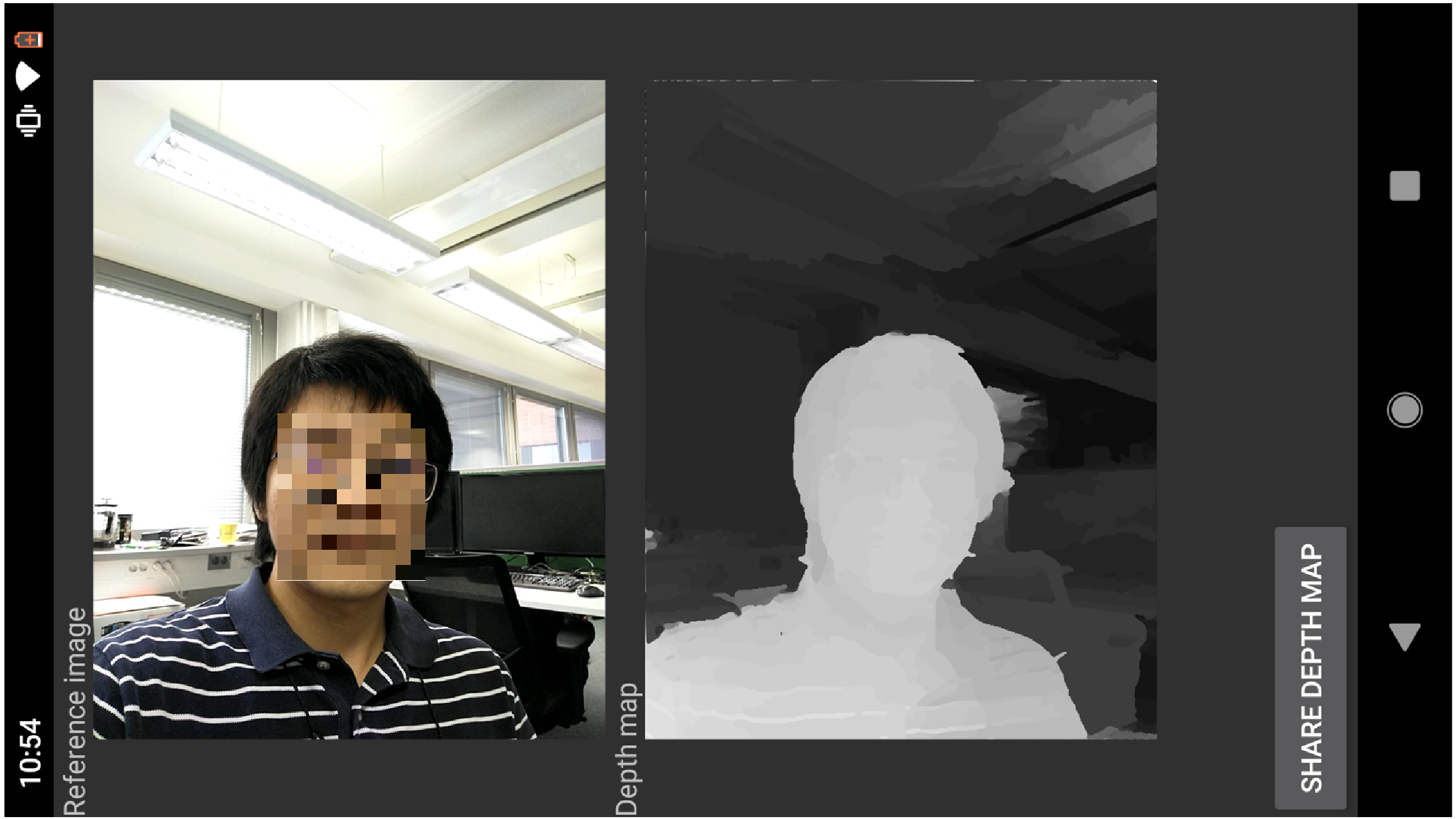}
	\end{center}

	\caption{Depth-map demo on "selfie" sequence for ANDROID mobile device using CPU-GPU optimized co-processing. 
	}
	\label{fig:onecol5}
\end{figure}

\subsection{Full Algorithm}
We summarize the proposed  DfSM using rank-1 initialization next in this section. This is a brief summary that incorporates the approach presented in ths paper and the DfSM described in \cite{Hyowon:2016}.


\begin{algorithm}
	\caption{DfSM with Rank-1 ({\bf \textit{iRank1} })}
	\label{alg:algorithm-label2}
	\noindent
	{\bf Input :} $p_{0j}$, $p_{ij}$, where $i=1\ldots n$ and $j=1\ldots m$ \\
	{\bf Output :} $K$, $k_1, k_2$, $R_i$, $c_i$, $\omega_j$, depth-map $\mathcal{D}$\\
	
	\vspace{0.1cm}
	{\bf Pre-processing :} \\
	\vspace{-0.5cm}
	\begin{itemize}
		\item  Estimate $\bar{p}_{ij}, \bar{p}_{0j}$ using $K$. The focal length is set to the larger value between image width and height. The principal point is set to the center of the image.	
	\end{itemize}
	\vspace{0.5cm}

	{\bf Bundle Adjustment using  Rank-1 initialization :} \\
	\vspace{-0.5cm}
	\begin{enumerate}
		
		\item  Estimate rotation $R_i$ between $\bar{p}_{0j} \leftrightarrow \bar{p}_{ij}$ using \cite{Kneip:2012}, 
		
		
		\item Rotate $\bar{p}_{i,j}$ to $\mathbf{p}_{ij}$ in the reference plane as eqn. (\ref{eqn:transh}),
		
		\item Create $\mathbb{M}$ matrix using $\mathbf{p}_{ij}$, and factorize as eqn. (\ref{err:rep})
		
		\item Refine the camera parameters and depth estimate using bundle adjustment \cite{Ceres}, \cite{Huber1992}. 
	\end{enumerate}

	{\bf Dense Matching :} \\	
	\vspace{-0.5cm}
	\begin{itemize}
		
		\item  Apply the dense matching proposed in \cite{Hyowon:2016} to determine the depth-map image $\mathcal{D}$.
	\end{itemize}

\end{algorithm}

				\begin{table*}[h!]
	\begin{center}
		\resizebox{.95\hsize}{!}{$
			\begin{tabular}{  | l  | l || c| c| c ||  c| c| c ||}
			\hline
			&	&  \multicolumn{3}{c|}{ {\bf \textit{10 frames} } } & \multicolumn{3}{c|}{ {\bf \textit{15 frames} } } \\  \cline{3-8} 	     
			{\bf \textit{ }}	 & &	& {\bf \textit{CPU-GPU}} & {\bf \textit{CPU-GPU}} & 	& {\bf \textit{CPU-GPU}} & {\bf \textit{CPU-GPU}} \\ 	 
			{\bf \textit{ Device}} & 	{\bf \textit{ Stages}} & {\bf \textit{CPU-only}} & {\bf \textit{iBA}} & {\bf \textit{iRank1}} & {\bf \textit{CPU-only}} & {\bf \textit{iBA}} & {\bf \textit{iRank1}} \\ 								
			\hline     \hline     
			\textit{ Bike1 }	&	\textit{Read input frame sequence }	 & 0.44 &  0.42 & 0.45 & 0.52 & 0.49 & 0.50 \\ \cline{2-8} 	
			&	{\bf\textit{Feature Extraction}} & 0.885& 0.539  & 0.539 & 1.375 & 0.771 & 0.771\\ 
			&	\hspace{1cm} \textit{Nb. features} & 1987  &  275 &  275   &  1941 & 258 & 258 \\ \cline{2-8}  
			&	{\bf \textit{Bundle Adjustment}	}  & {\bf 2.05} & {\bf 1.902}  & {\bf 0.824}   &  {\bf 2.303} & {\bf 2.315} & {\bf 0.571}  \\ 
			&	\hspace{1cm} {\bf  \textit{Nb. iteration}} & {\bf \textit{ NC-100}} &  {\bf \textit{ NC-100}} & {\bf  \textit{ C-20} }  &  {\bf \textit{ NC-100}} & {\bf \textit{C-100}} & {\bf \textit{C-15} } \\ \cline{2-8}  
			&	 \textit{Dense Matching}  	&  28.63 & 9.43 &  9.40   & 30.69   & 10.14 & 10.13  \\   \hline    \hline 
			\textit{ Bike2 }	&	\textit{Read input frame sequence }	 &  0.41 & 0.41   & 0.42  & 0.52 & 0.48 & 0.50 \\ \cline{2-8} 	
			&	{\bf\textit{Feature Extraction}}	  & 0.812 & 0.491   & 0.489 & 2.011 & 0.621 & 0.618\\ 
			&	\hspace{1cm} \textit{Nb. features} & 1958  &  293 & 293    & 1923  & 249  & 249  \\ \cline{2-8}  
			&	{\bf \textit{Bundle Adjustment}	}  & {\bf 2.13} & {\bf 1.851 }  & {\bf  0.803}     & {\bf 2.40  }  &  {\bf 2.12 }& {\bf 0.63} \\ 
			&	\hspace{1cm} {\bf \textit{Nb. iteration}} & {\bf \textit{ NC-100}} &  {\bf \textit{ NC-100}} & {\bf  \textit{C-23}}  & {\bf \textit{ NC-100}}  &  {\bf  \textit{C-95}} & {\bf \textit{C-10 }} \\ \cline{2-8}  
			&	 \textit{Dense Matching}  	& 28.48  & 9.14 &  9.16   &  30.12 & 10.29 & 10.24 \\   \hline   
			\end{tabular} $}
	\end{center}
	\caption{Table of total execution time in seconds (s) and convergence result using the Qualcomm snapdragon device on bike test video samples. {\bf \textit{ NC-100}} means no convergence at 100th iteration while \textit{C-20} means convergence at 20th iteration. }
	\label{tab:ddttdf2f1}
\end{table*}

\section{\uppercase{Experiments}}
To demonstrate the efficiency and practicality of the proposed implementation, we developed an interactive OpenCL ANDROID application that is shown in figure \ref{fig:onecol5}. The figure illustrates the reference image and the corresponding depth-map generated over all the 10 non-reference images.  Experiments were done with Qualcomm Snapdragon chipset containing Adreno 540  GPU, with CPU 4GB RAM and 8 cores. We implemented the proposed algorithm on a smartphone with GPU and CPU optimized co-processing in order to proper analyze the effectiveness of the proposed method directly on hand-held devices.

\subsection{Evaluation on Convergence}
We execute the algorithm on two test video clips "bike1", and "bike2". These test clips are full HD (1920 $\times$ 1080) video using  10 and 15 frames respectively. The process is initialized using the proposed grided feature extraction with a fixed grid size of 80 $\times$ 80, which provides 275 consistent features that is tracked all along the non-reference images. Without the grided feature extraction method, 1987 consistent features are expected to be tracked. Therefore, the proposed feature extraction allow approximately \textit{7}x reduction from the original features. The value of the grid size is optional and can be modified as seem fit by the user.  

\vspace{0.5cm}

Table \ref{tab:ddttdf2f1} provides a summarized execution time of the algorithms, and also justify the fast convergence of the proposed method. The {\bf \textit{CPU-only}} signifies the case when the DfSM algorithm is directly transfered to the mobile platform with some optimizations made but no grided feature extraction is done here. However, {\bf \textit{CPU-GPU}} is an ugraded version to the {\bf \textit{CPU-only}}, with GPU OpenCL acceleration on the dense matching using 128 depth-plane sweep, and the proposed grided feature extraction. In addition, {\bf \textit{iRank1}} represents the DfSM using rank-1 initialization proposed in this work while {\bf \textit{iBA}} represent the one without rank-1 initialization. 

This table exhibits two important informations; (1) the time complexity on mobile device measured in seconds and  (2) the convergence information. 	For the convergence part, we implore the user to focus on the {\bf \textit{Bundle Adjustment}} row. The {\bf \textit{iRank1}} method converges in approximately 25 iteration for the two bike examples using 10 images while {\bf \textit{iBA}} only partially converge at 50th iteration when more images were added to the acquisition.  Not only is the rank-1 initialization vital,  it is also fast and converge in about 3 iteration which is approximately 0.009s. In summary, this approach do not add any time complexity to the full algorithm and allow fast convergence of the bundle adjustment in as little as 10 iterations.


\vspace{0.5cm}

\begin{figure*}
	\begin{center}
		\includegraphics[height=10cm, width=15cm]{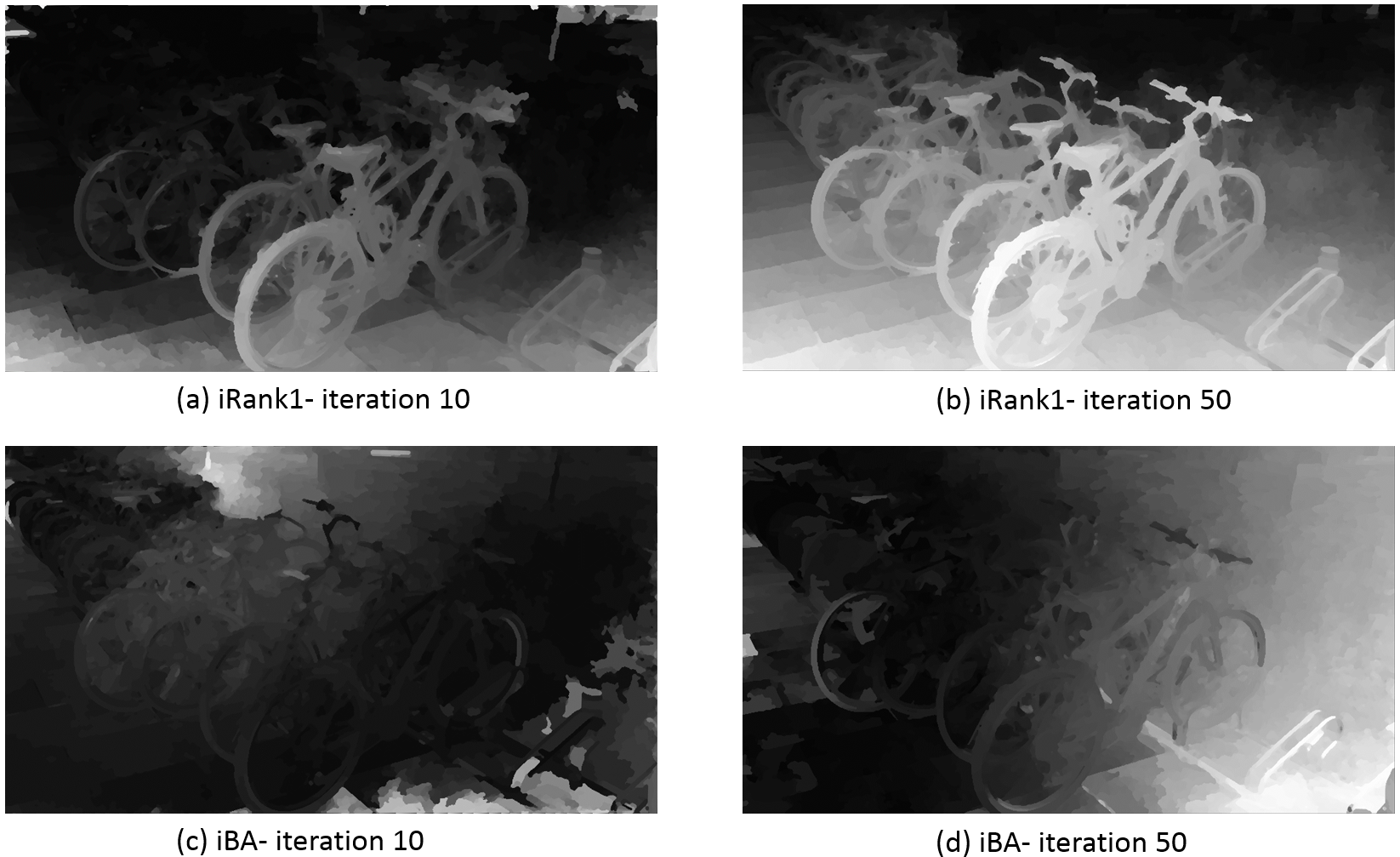}
	\end{center}
	\caption{ Subjective quality analysis of the depth-map generation between proposed {\bf \textit{iRank1}} and {\bf \textit{iBA}}, using 10 frames.
	}
	\label{fig:onecoasl32}
\end{figure*}

\begin{figure*}
	\begin{center}
		\includegraphics[height=10cm, width=15cm]{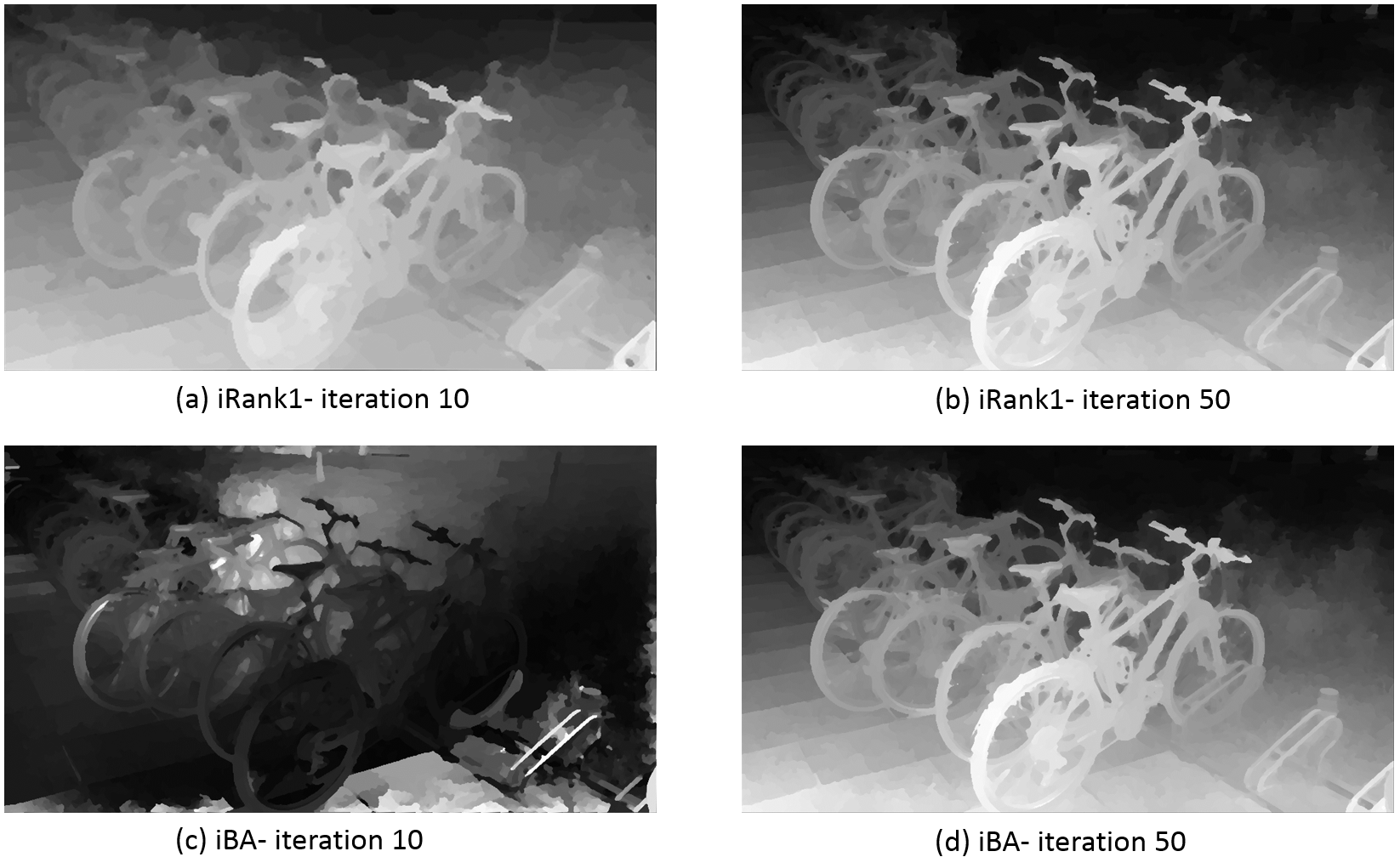}
	\end{center}
	\caption{ Subjective quality analysis of the depth-map generation between proposed {\bf \textit{iRank1}} and {\bf \textit{iBA}}, using 15 frames.
	}
	\label{fig:onecoash}
\end{figure*}

We made further test for subjective quality analysis of the proposed method on the earlier bike test sequence. Fig. \ref{fig:onecoasl32} illustrates the test with 10 frames while fig. \ref{fig:onecoash} illustrates the one with 15 frames. By analyzing the result in figure \ref{fig:onecoasl32}, one can see that {\bf \textit{iRank1}} method is already starting to converge at 10th iteration and finally converges at 50th iteration with pleasing depth-map result. However, for this same example, the {\bf \textit{iBA}} only start to converge at the 50th iteration. For the example in fig. \ref{fig:onecoash} using 15 frames, we can see that  the depth-map provided by {\bf \textit{iRank1}} at 10th iteration seems to have converged  while {\bf \textit{iBA}} only shows convergence at 50th iteration. With these experiments, we have shown the importance of rank-1 initialization under the DfSM algorithm.

\begin{table*}[h!]
	\begin{center}
		\resizebox{.95\hsize}{!}{$
			\begin{tabular}{  | l  | l    || c | c | c | c ||}
			\hline	 	     
			&	&	\multicolumn{2}{c|}{ {\bf \textit{Qualcomm Camera} } } &  \multicolumn{2}{c|}{ {\bf \textit{Hisilicon Camera} } } \\ \cline{3-6} 
			&	 	 &	{\bf \textit{Clip 1(5 videos)} } &	{\bf \textit{Clip 2(10 videos)} } &  	{\bf \textit{Clip 3(5 videos)} } &	{\bf \textit{Clip 4(10 videos)} } \\ 							
			\hline     \hline     
			\multirow{ 5}{*}{ \rot{90}{{\bf \textit{focal length} }} }	& {\bf \textit{ground-truth}}	&1360.21	& 1358.32  & 1503.71 &  1505.61 \\ 
			& {\bf \textit{Initial}}	& 1280.00	&  1280.00 & 1280.00 &  1280.00 \\ \cline{2-6} 
			& {\bf \textit{Min}}	& 1329.89	&  1341.84 & 1478.31&  1479.01 \\ 
			& {\bf \textit{Mean}}	& 1330.62	& 1335.16  &  1519.32 &  1503.78 \\ 
			& {\bf \textit{Max}}	& 1392.06	& 1371.69  & 1563.35 &  1531.32 \\ \hline \hline 		
			\multirow{ 4}{*}{ \rot{90}{{\bf \textit{Distortion} }} }	&{\bf \textit{Initial}}	& 4.25	&  4.08  & 5.72 & 5.63  \\	\cline{2-6} 
			&  {\bf \textit{Min}}	&0.51	& 0.39  & 0.08 & 0.09  \\ 
			& {\bf \textit{Mean}}	& 1.53	& 1.18  & 0.63 & 0.69  \\ 
			& {\bf \textit{Max}}	& 2.09 	&  1.34 & 1.09 &  1.15 \\  \hline  
			\end{tabular} $}
	\end{center}
	\caption{Evaluation in pixels unit for the estimated intrinsic camera parameters and radial distortion ones. We have tested 30 video clips in total acquired  by two cameras with different lens settings.}
	\label{tab:ddttdf2}
\end{table*}

\subsection{Evaluation on Self-Calibration}
As the bundle adjustment proposed in DfSM \cite{Hyowon:2016} is designed to self-calibrate the intrinsic camera parameters, we effect quantitative evaluation for the camera parameters obtained by the {\bf \textit{iRank1}} proposed in this work. For this experiment, we use two smartphones that contains Qualcomm snapdragon and Hisilicon Kirin chipsets. These two devices both have two different lens settings. The initial focal length is set to the largest value between the image width and height. We compare the estimated focal length and radial distortion against the ground truth. For measuring the distortion error, we generate a pixel grid and transform their coordinates using the estimated {\bf D}istorted-{\bf U}ndistorted image domain radial distortion function $\mathcal{F}$ explained in the original paper \cite{Hyowon:2016}. The transformed coordinates are again applied with the ground-truth {\bf U}ndistorted-{\bf D}istorted image domain model found in the camera pre-calibration. If the estimated $\mathcal{F}$ is reliable, these sequential transformation should be identity. The distortion error is measured in pixels using the mean of absolute distances. The result shows that the estimated parameters are close to the ground truth.

\vspace{-0.3cm}

Table \ref{tab:ddttdf2} shows the experimental result. We captured 15 videos each using camera located on the Qualcomm and Hisilicon devices. These cameras both have different lens settings, and the ground-truth camera parameters are acquired using the camera calibration toolbox \cite{Zhang:2000}.  For the qualcomm test, clip {\bf \textit{Clip 1(5 videos)} } and {\bf \textit{Clip 2(10 videos)} } make use of 5 and 10 videos, and each video containing 10 frames. These same procedure is repeated for Hisilicon device to create the test clips {\bf \textit{Clip 3(5 videos)} } and {\bf \textit{Clip 4(10 videos)}} respectively. In total, 30 video clips are used in this experiment. For the camera parameter, the focal length that is estimated are closer to the ground-truth. The mean distortion error for the Qualcomm is around 1.53 pixel while the initial parameter ($k_1, k_2 = 0$) gives an error of 4.25 pixel.



\section{\uppercase{Conclusion}}
The convergence of the the self-calibrating bundle adjustment needed to recover the camera parameters that is required for depth generation in the popular uncalibrated Depth from Small Motion algorithm \cite{Hyowon:2016} has not been well studied. Realistically, the convergence for the optimization procedure is not guaranteed even with the use of large number of frames (i.e approximately 30 images). In this work, we propose a new method  that incorporates the rank-1 factorization as a way to initialize the camera parameters and inverse depth points robustly. This approach allow fast convergence of the bundle adjustment procedure. 

\vspace{0.2cm}

The experimental results on a real mobile platform is presented. Compared with the state of art, our method can cope with a very small frame numbers to estimate the parameters required for good depth-map generation. After several optimizations with OpenCL GPU and CPU multi-threading procedures, the whole algorithm on the mobile device take approximately 10s for full HD resolution images. For the future work, we propose to investigate further the accuracy of the generated depth-map as compared to the state of art DfSM methods.

\vfill
\bibliographystyle{apalike}
{\small
\bibliography{egbib}}

\vfill
\end{document}